# B(eo)W(u)LF:
# Facilitating Recurrence Analysis on Multi-Level Language

*Version 1.0*
*Released August 12, 2013*

Alexandra Paxton and Rick Dale

Cognitive and Information Sciences
University of California, Merced
Merced, CA 95340
paxton.alexandra@gmail.com, http://www.alexandrapaxton.com

Discourse analysis may seek to characterize not only the overall composition of a given text or corpora but also the dynamic patterns within the data. Patterns of interest may occur at multiple levels, from character to sentence to corpus. Researchers may be interested in the way that sentence structures recur between participants or how affect words cluster in a single text. Recurrence analyses are an ideal tool for such investigations, but linguistic data must often be transformed prior to being analyzed.

This technical report introduces a data format called the *by-word long-form* or *B(eo)W(u)LF*. Inspired by the long-form data format required for mixed-effects modeling, B(eo)W(u)LF structures linguistic data into an expanded matrix encoding any number of researchers-specified markers. While we do not necessarily claim to be the first to use methods along these lines, we have created a series of tools utilizing Python and MATLAB to enable such discourse analyses.

We demonstrate this analysis on 319 lines of the Old English epic poem, *Beowulf*, translated into modern English (Appendix 1). At the end of this report, we provide the original text, scripts adapted to the text, and (for brevity's sake) a portion of the final result. The sample text is a single file, but if a corpus is saved in multiple individual files, these scripts can be modified with "if" statements to streamline the process.

## Text Preparation

These scripts require that text data be stored in a plain text file (.txt or .csv). If the corpus comprises separate text files, it's highly recommended that each file be transcribed or formatted identically. The data will be run through with a series of regular expressions in Python to quickly and automatically reformat the text (Appendix 2). Therefore, if individual files are formatted differently, the cleanup script will have to be tweaked for each format type.

The analyses to be performed will dictate the cleanup choices. In our example, we remove commas, colons, and semicolons, but we are interested in keeping the end-of-sentence punctuations (e.g., periods, question marks). We may choose to do some analyses on the separate cantos or lines of the poem, so we created indicators of each that will be utilized in the MATLAB script in the next section. After cleaning up the original text, the cleaned file is saved in a new file.

The cleaned file is then split by spaces, saving each word in a separate cell, and saved again. We format the data into a CSV format, but if the analyses require that commas remain in place, we recommend changing the Python script to save the output in a tab-delimited format. This file will then be read into the MATLAB script in the next phase.

## Initial Text Analysis

The next phase uses MATLAB to automatically calculate a number of target variables from the text (Appendix 3). These variables can be shaped according to researcher needs. For this example, we have chosen to create variables indicating canto number, poem line number, speech events, number of characters per word, and end-of-sentence markers. Of course, we could imagine a host of other variables that could also be chosen, like indicating when certain word(s) are used (e.g., name, location, experiment materials), exploring types of sentences (e.g., question marks, exclamation points), or tracking sentence length.

The code reads in the text produced by Python in the earlier phase and tracks canto and line number (using the indicators created earlier). We transform these indicators into separate variables and remove the indicators from the text. Once we also create the end-of-sentence, character number, and speech event variables, we save the data into a matrix (Appendix 4). Each line reports the values of each variable for a single word in the original text.

In anticipation of further linguistic analysis, we also save the text data in a .txt format with one word per line. In the following section, we model how to use the BWLF in conjunction with LIWC (Linguistic Inquiry and Word Count; Pennebaker, Booth, & Francis, 2007). However, these methods can be modified for other forms linguistic analyses as desired.

## Linguistic Analysis and Integration into BWLF Matrix

LIWC is a highly flexible and powerful tool that has been fairly widely used both in transcript- and text-based linguistic analyses (e.g., Niederhoffer & Pennebaker, 2002). In addition to providing pre-defined linguistic categories of both content and structure, LIWC allows researchers to specify their own categories of interest (e.g., a set of proper names). The text-only file that was produced in the previous section can be uploaded and analyzed by the LIWC program, using 1 newline as the delimiting option. Once the LIWC output has been saved, it can be integrated with the BLWF matrix we created earlier, again using MATLAB (Appendices 5-6).

## Concluding Remarks

B(eo)W(u)LF provides researchers interested in language with a data format that is well-suited for dynamic language analyses (e.g., recurrence analysis). Again, although we do not necessarily claim to be the first to have used data formats like this, we have here outlined a package of scripts for Python and MATLAB that can be easily modified to fit the specific needs of the project at hand. We believe that B(eo)W(u)LF provides an intuitively simple but analytically powerful data format particularly suited to analyzing language as it occurs and changes over a number of time scales.

## References

Niederhoffer, K. G., & Pennebaker, J. W. (2002). Linguistic style matching in social interaction. *Journal of Language and Social Psychology*, *21*(4), 337–360.

Pennebaker, J. W., Booth, R. J., & Francis, M. E. (2007). *Linguistic Inquiry and Word Count (LIWC): A computerized text analysis program*. Austin, TX: LIWC.net.

## Appendix 1:
## *Beowulf* Sample Text

Modern English translation of *Beowulf*. Taken from an e-text version by Robin Katsuya-Corbet (released into the public domain July 1993).

LO, praise of the prowess of people-kings
of spear-armed Danes, in days long sped,
we have heard, and what honor the athelings won!
Oft Scyld the Scefing from squadroned foes,
from many a tribe, the mead-bench tore,
awing the earls. Since erst he lay
friendless, a foundling, fate repaid him:
for he waxed under welkin, in wealth he throve,
till before him the folk, both far and near,
who house by the whale-path, heard his mandate,
gave him gifts: a good king he!
To him an heir was afterward born,
a son in his halls, whom heaven sent
to favor the folk, feeling their woe
that erst they had lacked an earl for leader
so long a while; the Lord endowed him,
the Wielder of Wonder, with world's renown.
Famed was this Beowulf: far flew the boast of him,
son of Scyld, in the Scandian lands.
So becomes it a youth to quit him well
with his father's friends, by fee and gift,
that to aid him, aged, in after days,
come warriors willing, should war draw nigh,
liegemen loyal: by lauded deeds
shall an earl have honor in every clan.
Forth he fared at the fated moment,
sturdy Scyld to the shelter of God.
Then they bore him over to ocean's billow,
loving clansmen, as late he charged them,
while wielded words the winsome Scyld,
the leader beloved who long had ruled....
In the roadstead rocked a ring-dight vessel,
ice-flecked, outbound, atheling's barge:
there laid they down their darling lord
on the breast of the boat, the breaker-of-rings,
by the mast the mighty one. Many a treasure
fetched from far was freighted with him.
No ship have I known so nobly dight
with weapons of war and weeds of battle,
with breastplate and blade: on his bosom lay
a heaped hoard that hence should go
far o'er the flood with him floating away.
No less these loaded the lordly gifts,
thanes' huge treasure, than those had done
who in former time forth had sent him
sole on the seas, a suckling child.
High o'er his head they hoist the standard,
a gold-wove banner; let billows take him,
gave him to ocean. Grave were their spirits,
mournful their mood. No man is able
to say in sooth, no son of the halls,
no hero 'neath heaven, - who harbored that freight!
Now Beowulf bode in the burg of the Scyldings,
leader beloved, and long he ruled
in fame with all folk, since his father had gone
away from the world, till awoke an heir,
haughty Healfdene, who held through life,
sage and sturdy, the Scyldings glad.
Then, one after one, there woke to him,
to the chieftain of clansmen, children four:
Heorogar, then Hrothgar, then Halga brave;
and I heard that - was -'s queen,
the Heathoscylfing's helpmate dear.
To Hrothgar was given such glory of war,
such honor of combat, that all his kin
obeyed him gladly till great grew his band
of youthful comrades. It came in his mind
to bid his henchmen a hall uprear,
a master mead-house, mightier far
than ever was seen by the sons of earth,
and within it, then, to old and young
he would all allot that the Lord had sent him,
save only the land and the lives of his men.
Wide, I heard, was the work commanded,
for many a tribe this mid-earth round,
to fashion the folkstead. It fell, as he ordered,
in rapid achievement that ready it stood there,
of halls the noblest: Heorot he named it
whose message had might in many a land.
Not reckless of promise, the rings he dealt,
treasure at banquet: there towered the hall,
high, gabled wide, the hot surge waiting
of furious flame. Nor far was that day
when father and son-in-law stood in feud
for warfare and hatred that woke again.
With envy and anger an evil spirit
endured the dole in his dark abode,
that he heard each day the din of revel
high in the hall: there harps rang out,
clear song of the singer. He sang who knew tales of the early time of man,
how the Almighty made the earth,
fairest fields enfolded by water,
set, triumphant, sun and moon
for a light to lighten the land-dwellers,
and braided bright the breast of earth
with limbs and leaves, made life for all
of mortal beings that breathe and move.
So lived the clansmen in cheer and revel
a winsome life, till one began
to fashion evils, that field of hell.
Grendel this monster grim was called,
march-riever mighty, in moorland living,
in fen and fastness; fief of the giants
the hapless wight a while had kept
since the Creator his exile doomed.
On kin of Cain was the killing avenged
by sovran God for slaughtered Abel.
Ill fared his feud, and far was he driven,
for the slaughter's sake, from sight of men.
Of Cain awoke all that woful breed,
Etins and elves and evil-spirits,
as well as the giants that warred with God
weary while: but their wage was paid them!
WENT he forth to find at fall of night
that haughty house, and heed wherever
the Ring-Danes, outrevelled, to rest had gone.
Found within it the atheling band
asleep after feasting and fearless of sorrow,
of human hardship. Unhallowed wight,
grim and greedy, he grasped betimes,
wrathful, reckless, from resting-places,
thirty of the thanes, and thence he rushed
fain of his fell spoil, faring homeward,
laden with slaughter, his lair to seek.
Then at the dawning, as day was breaking,
the might of Grendel to men was known;
then after wassail was wail uplifted,
loud moan in the morn. The mighty chief,
atheling excellent, unblithe sat,
labored in woe for the loss of his thanes,
when once had been traced the trail of the fiend,
spirit accurst: too cruel that sorrow,
too long, too loathsome. Not late the respite;
with night returning, anew began
ruthless murder; he recked no whit,
firm in his guilt, of the feud and crime.
They were easy to find who elsewhere sought
in room remote their rest at night,
bed in the bowers, when that bale was shown,
was seen in sooth, with surest token, -
the hall-thane's hate. Such held themselves
far and fast who the fiend outran!
Thus ruled unrighteous and raged his fill
one against all; until empty stood
that lordly building, and long it bode so.
Twelve years' tide the trouble he bore,
sovran of Scyldings, sorrows in plenty,
boundless cares. There came unhidden
tidings true to the tribes of men,
in sorrowful songs, how ceaselessly Grendel
harassed Hrothgar, what hate he bore him,
what murder and massacre, many a year,
feud unfading, - refused consent
to deal with any of Daneland's earls,
make pact of peace, or compound for gold:
still less did the wise men ween to get
great fee for the feud from his fiendish hands.
But the evil one ambushed old and young
death-shadow dark, and dogged them still,
lured, or lurked in the livelong night
of misty moorlands: men may say not
where the haunts of these Hell-Runes be.
Such heaping of horrors the hater of men,
lonely roamer, wrought unceasing,
harassings heavy. O'er Heorot he lorded,
gold-bright hall, in gloomy nights;
and ne'er could the prince approach his throne,
- 'twas judgment of God, - or have joy in his hall.
Sore was the sorrow to Scyldings'-friend,
heart-rending misery. Many nobles
sat assembled, and searched out counsel
how it were best for bold-hearted men
against harassing terror to try their hand.
Whiles they vowed in their heathen fanes
altar-offerings, asked with words
that the slayer-of-souls would succor give them
for the pain of their people. Their practice this,
their heathen hope; 'twas Hell they thought of
in mood of their mind. Almighty they knew not,
Doomsman of Deeds and dreadful Lord,
nor Heaven's-Helmet heeded they ever,
Wielder-of-Wonder. - Woe for that man
who in harm and hatred hales his soul
to fiery embraces; - nor favor nor change
awaits he ever. But well for him
that after death-day may draw to his Lord,
and friendship find in the Father's arms!
THUS seethed unceasing the son of Healfdene
with the woe of these days; not wisest men
assuaged his sorrow; too sore the anguish,
loathly and long, that lay on his folk,
most baneful of burdens and bales of the night.

This heard in his home Hygelac's thane,
great among Geats, of Grendel's doings.
He was the mightiest man of valor
in that same day of this our life,
stalwart and stately. A stout wave-walker
he bade make ready. Yon battle-king, said he,
far o'er the swan-road he fain would seek,
the noble monarch who needed men!
The prince's journey by prudent folk
was little blamed, though they loved him dear;
they whetted the hero, and hailed good omens.
And now the bold one from bands of Geats
comrades chose, the keenest of warriors
e'er he could find; with fourteen men
the sea-wood he sought, and, sailor proved,
led them on to the land's confines.
Time had now flown; afloat was the ship,
boat under bluff. On board they climbed,
warriors ready; waves were churning
sea with sand; the sailors bore
on the breast of the bark their bright array,
their mail and weapons: the men pushed off,
on its willing way, the well-braced craft.
Then moved o'er the waters by might of the wind
that bark like a bird with breast of foam,
till in season due, on the second day,
the curved prow such course had run
that sailors now could see the land,
sea-cliffs shining, steep high hills,
headlands broad. Their haven was found,
their journey ended. Up then quickly
the Weders' clansmen climbed ashore,
anchored their sea-wood, with armor clashing
and gear of battle: God they thanked
for passing in peace o'er the paths of the sea.
Now saw from the cliff a Scylding clansman,
a warden that watched the water-side,
how they bore o'er the gangway glittering shields,
war-gear in readiness; wonder seized him
to know what manner of men they were.
Straight to the strand his steed he rode,
Hrothgar's henchman; with hand of might
he shook his spear, and spake in parley.
"Who are ye, then, ye armed men,
mailed folk, that yon mighty vessel
have urged thus over the ocean ways,
here o'er the waters? A warden I,
sentinel set o'er the sea-march here,
lest any foe to the folk of Danes
with harrying fleet should harm the land.
No aliens ever at ease thus bore them,
linden-wielders: yet word-of-leave
clearly ye lack from clansmen here,
my folk's agreement. - A greater ne'er saw I
of warriors in world than is one of you, -
yon hero in harness! No henchman he
worthied by weapons, if witness his features,
his peerless presence! I pray you, though, tell
your folk and home, lest hence ye fare
suspect to wander your way as spies
in Danish land. Now, dwellers afar,
ocean-travellers, take from me
simple advice: the sooner the better
I hear of the country whence ye came."
To him the stateliest spake in answer;
the warriors' leader his word-hoard unlocked:-
"We are by kin of the clan of Geats,
and Hygelac's own hearth-fellows we.
To folk afar was my father known,
noble atheling, Ecgtheow named.
Full of winters, he fared away
aged from earth; he is honored still
through width of the world by wise men all.
To thy lord and liege in loyal mood
we hasten hither, to Healfdene's son,
people-protector: be pleased to advise us!
To that mighty-one come we on mickle errand,
to the lord of the Danes; nor deem I right
that aught be hidden. We hear - thou knowest
if sooth it is - the saying of men,
that amid the Scyldings a scathing monster,
dark ill-doer, in dusky nights
shows terrific his rage unmatched,
hatred and murder. To Hrothgar I
in greatness of soul would succor bring,
so the Wise-and-Brave may worst his foes, -
if ever the end of ills is fated,
of cruel contest, if cure shall follow,
and the boiling care-waves cooler grow;
else ever afterward anguish-days
he shall suffer in sorrow while stands in place
high on its hill that house unpeered!"
Astride his steed, the strand-ward answered,
clansman unquailing: "The keen-souled thane
must be skilled to sever and sunder duly
words and works, if he well intends.
I gather, this band is graciously bent
to the Scyldings' master. March, then, bearing
weapons and weeds the way I show you.
I will bid my men your boat meanwhile
to guard for fear lest foemen come, -
your new-tarred ship by shore of ocean
faithfully watching till once again
it waft o'er the waters those well-loved thanes,

- winding-neck'd wood, - to Weders' bounds,
heroes such as the hest of fate
shall succor and save from the shock of war."
They bent them to march, - the boat lay still,
fettered by cable and fast at anchor,
broad-bosomed ship. - Then shone the boars
over the cheek-guard; chased with gold,
keen and gleaming, guard it kept
o'er the man of war, as marched along
heroes in haste, till the hall they saw,
broad of gable and bright with gold:
that was the fairest, 'mid folk of earth,
of houses 'neath heaven, where Hrothgar lived,
and the gleam of it lightened o'er lands afar.
The sturdy shieldsman showed that bright
burg-of-the-boldest; bade them go
straightway thither; his steed then turned,
hardy hero, and hailed them thus:-
"Tis time that I fare from you. Father Almighty
in grace and mercy guard you well,
safe in your seekings. Seaward I go,
'gainst hostile warriors hold my watch."

# Appendix 2:
# Python Source Code for Data Preparation

This code is also available for upon request.

```
###############
# B(eo)W(u)LF Code:  By-Word Long-Form Data Preparation

# This code reads in a sample text file, reformats it, and exports it
# to a new file before further formatting and analysis.

# Written by: Alexandra Paxton, University of California, Merced
# Date last modified: June 16, 2013
###############

# coding:utf-8
import os,re,unicodedata,shlex,glob

# read in text file
os.chdir('~/bwlfTechReport/')
beowulfText = open('beowulfTextSnippet.txt','r') # open file
beowulf = beowulfText.read() # read in text

# start the cleanup (change according to text)
beowulf = re.sub('\r','\n',beowulf) # ensure all newlines are identical
beowulf = re.sub('(^|\n)([A-Z]{2,}( |,))','\n\n[canto]\n\\2',beowulf) #
create canto indicator
beowulf = re.sub(':|;|(\-)|,',' ',beowulf) # remove non-target
punctuation
beowulf = re.sub('(\.){3,}','.',beowulf)
beowulf = re.sub(' {1,}',' ',beowulf) # remove redundant spaces

# convert all text to lower case
beowulf = beowulf.lower()

# split by spaces
beowulf = re.sub('\n(?=([A-Z]|[a-z]|\"))',' [line] ',beowulf) # create
line indicator
beowulf = re.sub('\n',' ',beowulf) # convert newline to space
beowulf = re.split(' +',beowulf) # split file by space
beowulf = str(beowulf) # convert to a string
beowulf = re.sub('((\')|("))\, ((\')|("))',',',beowulf) # remove
extraneous quotations
beowulf = re.sub('\,{2}',',',beowulf) # remove empty cells
beowulf = re.sub('(\[\'(,)?)|(\'\])','',beowulf) # remove extraneous
brackets

# close file and print new file
beowulfText.close()
cleaned = file('beowulfCleaned.csv','w')
cleaned.write(beowulf)
cleaned.close()
```

## Appendix 3:
## MATLAB Source Code for Initial Text Analysis

This code is also available for upon request.

```
%% B(eo)W(u)LF Code: By-Word Long-Form Initial Text Analysis

% This code reads in the cleaned file produced by the Python script,
% computes a series of automatic variables from the texts, then outputs
% two files: a matrix with the automatic variables and a text file
% suitable for other linguistic analyses (e.g., LIWC).

% Written by: Alexandra Paxton, University of California, Merced
% Date last modified: June 17, 2013

%%

% preliminaries
clear
cd('./bwlfTechReport');

% read in cleaned text file
bText = fopen('beowulfCleaned.csv');
bRead = textscan(bText,'%s','EndOfLine','\n','delimiter',',');
bRead = bRead{1,1};
disp('Cleaned Text File Loaded.')

% separate and renumber cantos
canto = regexp(bRead,'\[canto\]');
cantoCount = 0;
trashLines = [];
for cantos = 1:length(bRead)
    if cellfun(@isempty,canto(cantos))==0
        cantoCount = cantoCount + 1;
        trashLines = [trashLines cantos];
    else
        cantoTrack(cantos,1) = cantoCount;
    end
end
disp('Cantos Numbered.')

% separate and renumber lines
pLine = regexp(bRead,'\[line\]');
pLineCount = 0;
for pLines = 1:length(bRead)
    if cellfun(@isempty,pLine(pLines))==0
        pLineCount = pLineCount + 1;
        trashLines = [trashLines pLines];
    else
        pLineTrack(pLines,1) = pLineCount;
    end
end
disp('Poem Lines Numbered.')

% remove indicator lines from text and indicator matrices
bRead(trashLines) = [];
```

```
cantoTrack(trashLines) = [];
pLineTrack(trashLines) = [];

% find end-of-sentence lines
ends = regexp(bRead,'.*[\.|?\!]');
endsNone = cellfun(@isempty,ends);
disp('Sentences Isolated.')

% track speech lines
speech = regexp(bRead,'.*\"');
speechTrack = [];
tempStore = [];
speechEvent = 1;
for speaking = 1:length(bRead)
    if cellfun(@isempty,speech(speaking))==0
        tempStore = [tempStore speaking];
        if length(tempStore)==2
            speechTrack(speechEvent,1:2) = tempStore;
            tempStore = [];
            speechEvent = speechEvent + 1;
        end
    end
end
sTrack = 1;
for speechMark = 1:length(bRead)
    if speechTrack(sTrack,1) <= speechMark && speechMark<=
speechTrack(sTrack,2)
        speech{speechMark} = 1;
        if speechMark == speechTrack(sTrack,2)
            sTrack = sTrack + 1;
        end
    end
end
speechNone = cellfun(@isempty,speech);
disp('Speech Isolated.')

% create initial by-word long-form matrix
for i = 1:length(bRead)
    % track sentence ends
    if endsNone(i) == 0
        eos = 1; % indicates end of sentence
        charNum = length(char(bRead{i}))-1; % tracks current word
length, minus the punctuation
    else
        eos = 0; % indicates not end of sentence
        charNum = length(char(bRead{i})); % tracks current word length,
minus the punctuation
    end

    % track speech events
    if speechNone(i) == 0
        sp = 1; % indicates speech event
        if regexp(bRead{i},'.*\"')==1
            charNum = charNum - 1; % subtrack quotation mark from
character count
        end
    else
```

```
        sp = 0; % indicates no speech event
    end

    % store everything in matrix
    beowulfMat(i,:) =
{int2str(cantoTrack(i)),int2str(pLineTrack(i)),bRead{i},int2str(charNum
),int2str(sp),int2str(eos)};
    if mod(i,500)==0;
        disp(['Line ' int2str(i) ' of ' int2str(length(bRead)) '
Recorded.'])
    end
end

% save workspace
save beowulfBWLF.mat
disp('MATLAB Workspace Saved.')

% print transcript for LIWC
textFileName = ('bwlfTextAnalysisPrep.txt');
textLine = beowulfMat(:,3);
textFile = fopen(textFileName,'w');
for word = 1:length(textLine)
    fprintf(textFile,'%s\n',textLine{word});
end

% print matrix
matrixFile = ('beowulfBWLFMatrix.csv');
matOut = fopen(matrixFile,'w');
header_names = {'canto';'line';'word';'charnum';'speech';'eos'};
fprintf(matOut,'%s,%s,%s,%s,%s,%s\n',header_names{:});
for beoLine = 1:size(beowulfMat,1)
    fprintf(matOut,'%s,%s,%s,%s,%s,%s\n',beowulfMat{beoLine,:});
end
disp('Matrix Output Complete.')
save beowulfBLWF.mat

% close file
fclose(matOut);
disp('Processing Complete.')
```

## Appendix 4:
## 39 Lines of BWLF Output

| *canto* | *line* | *word* | *charnum* | *speech* | *eos* |
|---|---|---|---|---|---|
| 1 | 1 | lo | 2 | 0 | 0 |
| 1 | 1 | praise | 6 | 0 | 0 |
| 1 | 1 | of | 2 | 0 | 0 |
| 1 | 1 | the | 3 | 0 | 0 |
| 1 | 1 | prowess | 7 | 0 | 0 |
| 1 | 1 | of | 2 | 0 | 0 |
| 1 | 1 | people | 6 | 0 | 0 |
| 1 | 1 | kings | 5 | 0 | 0 |
| 1 | 2 | of | 2 | 0 | 0 |
| 1 | 2 | spear | 5 | 0 | 0 |
| 1 | 2 | armed | 5 | 0 | 0 |
| 1 | 2 | danes | 5 | 0 | 0 |
| 1 | 2 | in | 2 | 0 | 0 |
| 1 | 2 | days | 4 | 0 | 0 |
| 1 | 2 | long | 4 | 0 | 0 |
| 1 | 2 | sped | 4 | 0 | 0 |
| 1 | 3 | we | 2 | 0 | 0 |
| 1 | 3 | have | 4 | 0 | 0 |
| 1 | 3 | heard | 5 | 0 | 0 |
| 1 | 3 | and | 3 | 0 | 0 |
| 1 | 3 | what | 4 | 0 | 0 |
| 1 | 3 | honor | 5 | 0 | 0 |
| 1 | 3 | the | 3 | 0 | 0 |
| 1 | 3 | athelings | 9 | 0 | 0 |
| 1 | 3 | won! | 3 | 0 | 1 |
| 1 | 4 | oft | 3 | 0 | 0 |
| 1 | 4 | scyld | 5 | 0 | 0 |
| 1 | 4 | the | 3 | 0 | 0 |
| 1 | 4 | scefing | 7 | 0 | 0 |
| 1 | 4 | from | 4 | 0 | 0 |
| 1 | 4 | squadroned | 10 | 0 | 0 |
| 1 | 4 | foes | 4 | 0 | 0 |
| 1 | 5 | from | 4 | 0 | 0 |
| 1 | 5 | many | 4 | 0 | 0 |
| 1 | 5 | a | 1 | 0 | 0 |
| 1 | 5 | tribe | 5 | 0 | 0 |
| 1 | 5 | the | 3 | 0 | 0 |
| 1 | 5 | mead | 4 | 0 | 0 |

## Appendix 5:
## MATLAB Source Code for Data Integration

This code is also available for upon request.

```
%% B(eo)W(u)LF Code: By-Word Long-Form Data Integration

% This code combines the BWLF matrix created with the previous
% MATLAB script with LIWC output to create a single file.

% Written by: Alexandra Paxton, University of California, Merced
% Date last modified: June 17, 2013
%%

% preliminaries
clear
cd('./bwlfTechReport');

% import LIWC data
liwcData = importdata('beowulfLIWC.txt');
disp('LIWC Data Imported.')

% create variables named for each header
headers = liwcData.textdata(1,:);
Filename = liwcData.textdata(2:end,1);
for i = 2:(length(headers))
    eval([headers{i} ' = liwcData.data(:,' int2str(i-1) ');']);
end
disp('LIWC Category Variable Headers Created.')

% import workspace
load beowulfBWLF.mat

% create output file for integrated matrx
matrixFile = ('beowulfBwlfLiwc.csv');
matOut = fopen(matrixFile,'w');

% print headers
fprintf(matOut,'%s,',header_names{:});
fprintf(matOut,'%s,',headers{1:length(headers)-1});
fprintf(matOut,'%s\n',headers{length(headers)});

% print to file
for thisLine = 1:length(bRead)
    fprintf(matOut,'%s,',beowulfMat{thisLine,1:6});
    fprintf(matOut,'%s,',Filename{thisLine});
    fprintf(matOut,'%d,',liwcData.data(thisLine,1:(length(headers)-
2)));
    fprintf(matOut,'%d\n',liwcData.data(thisLine,length(headers)-1));
    if mod(thisLine,500)==0;
        disp(['Line ' int2str(thisLine) ' Recorded.'])
    end
end
fclose(matOut);
disp('Processing Complete.')
```

## Appendix 6:
## 39 Lines of BWLF Output with (Partial) Integrated LIWC Analyses

| canto | line | word | charnum | speech | eos | Seg | WC | WPS | Sixltr | Dic |
|---|---|---|---|---|---|---|---|---|---|---|
| 1 | 1 | lo | 2 | 0 | 0 | 1 | 1 | 1 | 0 | 0 |
| 1 | 1 | praise | 6 | 0 | 0 | 2 | 1 | 1 | 0 | 100 |
| 1 | 1 | of | 2 | 0 | 0 | 3 | 1 | 1 | 0 | 100 |
| 1 | 1 | the | 3 | 0 | 0 | 4 | 1 | 1 | 0 | 100 |
| 1 | 1 | prowess | 7 | 0 | 0 | 5 | 1 | 1 | 100 | 0 |
| 1 | 1 | of | 2 | 0 | 0 | 6 | 1 | 1 | 0 | 100 |
| 1 | 1 | people | 6 | 0 | 0 | 7 | 1 | 1 | 0 | 100 |
| 1 | 1 | kings | 5 | 0 | 0 | 8 | 1 | 1 | 0 | 100 |
| 1 | 2 | of | 2 | 0 | 0 | 9 | 1 | 1 | 0 | 100 |
| 1 | 2 | spear | 5 | 0 | 0 | 10 | 1 | 1 | 0 | 0 |
| 1 | 2 | armed | 5 | 0 | 0 | 11 | 1 | 1 | 0 | 0 |
| 1 | 2 | danes | 5 | 0 | 0 | 12 | 1 | 1 | 0 | 0 |
| 1 | 2 | in | 2 | 0 | 0 | 13 | 1 | 1 | 0 | 100 |
| 1 | 2 | days | 4 | 0 | 0 | 14 | 1 | 1 | 0 | 100 |
| 1 | 2 | long | 4 | 0 | 0 | 15 | 1 | 1 | 0 | 100 |
| 1 | 2 | sped | 4 | 0 | 0 | 16 | 1 | 1 | 0 | 100 |
| 1 | 3 | we | 2 | 0 | 0 | 17 | 1 | 1 | 0 | 100 |
| 1 | 3 | have | 4 | 0 | 0 | 18 | 1 | 1 | 0 | 100 |
| 1 | 3 | heard | 5 | 0 | 0 | 19 | 1 | 1 | 0 | 100 |
| 1 | 3 | and | 3 | 0 | 0 | 20 | 1 | 1 | 0 | 100 |
| 1 | 3 | what | 4 | 0 | 0 | 21 | 1 | 1 | 0 | 100 |
| 1 | 3 | honor | 5 | 0 | 0 | 22 | 1 | 1 | 0 | 100 |
| 1 | 3 | the | 3 | 0 | 0 | 23 | 1 | 1 | 0 | 100 |
| 1 | 3 | athelings | 9 | 0 | 0 | 24 | 1 | 1 | 100 | 0 |
| 1 | 3 | won! | 3 | 0 | 1 | 25 | 1 | 1 | 0 | 100 |
| 1 | 4 | oft | 3 | 0 | 0 | 26 | 1 | 1 | 0 | 0 |
| 1 | 4 | scyld | 5 | 0 | 0 | 27 | 1 | 1 | 0 | 0 |
| 1 | 4 | the | 3 | 0 | 0 | 28 | 1 | 1 | 0 | 100 |
| 1 | 4 | scefing | 7 | 0 | 0 | 29 | 1 | 1 | 100 | 0 |
| 1 | 4 | from | 4 | 0 | 0 | 30 | 1 | 1 | 0 | 100 |
| 1 | 4 | squadroned | 10 | 0 | 0 | 31 | 1 | 1 | 100 | 0 |
| 1 | 4 | foes | 4 | 0 | 0 | 32 | 1 | 1 | 0 | 100 |
| 1 | 5 | from | 4 | 0 | 0 | 33 | 1 | 1 | 0 | 100 |
| 1 | 5 | many | 4 | 0 | 0 | 34 | 1 | 1 | 0 | 100 |
| 1 | 5 | a | 1 | 0 | 0 | 35 | 1 | 1 | 0 | 100 |
| 1 | 5 | tribe | 5 | 0 | 0 | 36 | 1 | 1 | 0 | 0 |
| 1 | 5 | the | 3 | 0 | 0 | 37 | 1 | 1 | 0 | 100 |
| 1 | 5 | mead | 4 | 0 | 0 | 38 | 1 | 1 | 0 | 0 |